\documentclass[journal]{IEEEtran}

\usepackage{xcolor,soul,framed} 

\colorlet{shadecolor}{yellow}
\usepackage{graphicx}
\graphicspath{{../pdf/}{../jpeg/}}
\DeclareGraphicsExtensions{.pdf,.jpeg,.png}

\usepackage[cmex10]{amsmath}
\usepackage{array}
\usepackage{mdwmath}
\usepackage{mdwtab}
\usepackage{eqparbox}
\usepackage{url}

\begin{document}
	\bstctlcite{IEEEexample:BSTcontrol}
	\title{Teleoperated Haircutting with Delay Compensation and Passivity-Based Control for Unilateral Internet Teleoperation under Real Communication Uncertainties} 

\author{
	\begin{tabular}{c}
		Viridiana Luna-Zempoalteca\textsuperscript{1},
		Zhendai Huang\textsuperscript{2},
		Rosalba Galvan-Guerra\textsuperscript{1},
		Antonio Rosales\textsuperscript{3},
		Xiaoli Liu\textsuperscript{4},\\
		and Shuai Li (Steven)\textsuperscript{2}
		\\[2mm]
		\parbox{0.95\textwidth}{
			\centering
			\footnotesize
			
			\textsuperscript{1} Instituto Politécnico Nacional, Mexico.
			
			\textsuperscript{2} University of Oulu, Oulu, Finland.
			
			\textsuperscript{3} VTT Technical Research Centre of Finland, Oulu, Finland.
			
			\textsuperscript{4} University of Helsinki, Finland.
		}
	\end{tabular}
}

	\markboth{	}{ \MakeLowercase{\textit{}}: Title}

	\maketitle

	
	\begin{abstract}
		Robotic haircutting requires accurate motion execution and reliable communication to safely reproduce the operator's commands.
		Internet-based teleoperation is a key enabling technology to remotely perform precise manipulation tasks such as haircutting. However, communication uncertainties, including variable transmission delays and packet jitter, degrade synchronization between the master and slave systems and affect tracking performance. This paper presents a unilateral Internet teleoperation framework for robotic haircutting that takes into account experimentally measured communication uncertainties. Three delay-compensation techniques, namely Zero-Order Hold (ZOH), First-Order Hold (FOH), and Low-Pass Filtering (LPF), are evaluated together with a classical proportional-derivative (PD) controller and a passivity-based synchronization controller. Unlike conventional studies based on synthetic delay models, the proposed framework is validated using communication data acquired from a real cloud-based teleoperation platform, where both packet delay and effective delay are characterized. The results demonstrate that combining delay-compensation techniques with passivity-based synchronization improves the robustness of unilateral Internet teleoperation under realistic communication uncertainties, providing a reliable control framework for robotic haircutting applications.
		
	\end{abstract}

	\begin{IEEEkeywords}
		Internet teleoperation, passivity-based control, passive synchronization, communication delay, jitter compensation, cloud robotics, ArUco tracking.
	\end{IEEEkeywords}
	
	%
	\IEEEpeerreviewmaketitle
	
	\section{Introduction}
	Haircutting robots have recently emerged \cite{Li_2026} as a promising application of service robotics, driven by advances in robotic manipulation, artificial intelligence \cite{BOGUE2020}, computer vision \cite{Bejczy1987}, and cloud robotics \cite{Zhendai2026}. Unlike conventional industrial robotic tasks, robotic haircutting requires accurate motion control, continuous visual perception, and safe physical interaction with the human head. These stringent requirements make robustness, motion smoothness, and communication reliability essential for practical deployment \cite{Shuai2025}. Despite these advances, achieving fully autonomous robotic haircutting remains challenging due to the complexity of real-time perception, dexterous manipulation, and safety-critical human–robot interaction.
	
	Teleoperation offers an attractive intermediate solution that bridges the gap between manual haircutting and fully autonomous operation by combining human expertise with robotic precision \cite{Sanders2009}. Through Internet-based communication, a human operator can remotely supervise or directly perform haircutting tasks while the robot accurately executes the desired motions \cite{Khan2026}. Consequently, teleoperation has become an enabling technology not only for robotic haircutting but also for numerous robotic applications including telesurgery, industrial maintenance, underwater inspection, hazardous-environment intervention, and collaborative robotics \cite{HOKAYEM2006,Sharifi2017}. 
	
	Despite these advantages, Internet-based teleoperation is strongly influenced by communication uncertainties \cite{Chopra2003}. Variable transmission delays, packet losses, communication jitter, and asynchronous packet arrivals deteriorate synchronization between master and slave systems, reducing the accuracy of the tracking and potentially compromising the stability of the system \cite{Anderson1992}. As cloud robotics continues to expand, these communication effects have become a principal challenge in the design of reliable teleoperation systems \cite{Rosas2022}.
	
	Several approaches have been proposed to mitigate communication-induced uncertainties. Reference reconstruction techniques such as Zero-Order Hold (ZOH) \cite{Zheng2007}, First-Order Hold (FOH) \cite{Leonard1999}, and Low-Pass Filtering (LPF) \cite{Proakis2006} are commonly employed to generate continuous motion commands despite irregular packet arrivals. These techniques exhibit different trade-offs among tracking accuracy, smoothness, and computational complexity, yet their comparative performance under real cloud-network conditions remains underexplored.
	
	In addition to signal reconstruction, passivity-based control has become \cite{Sun2017} one of the most widely adopted approaches to guaranteeing stable teleoperation under communication delays \cite{NUNO2008}. Classical scattering-based methods preserve passivity by transforming the communication channel into a passive subsystem, whereas synchronization-based approaches simplify the implementation while maintaining delay-independent stability properties \cite{Chopra2005}. These methods have established passivity as one of the most robust theoretical frameworks for networked teleoperation \cite{NUNO2011}.
	
	Most existing studies evaluate teleoperation controllers using simulated communication channels or idealized constant-delay models \cite{Chicaiza2024}. Although significant progress has been achieved, relatively few works experimentally investigate unilateral cloud-based teleoperation operating under real Internet communication uncertainties while 
	systematically analyzing the influence of different delay-compensation techniques on controller performance.
	
	To address this research gap, this paper presents a unilateral cloud-based teleoperation framework for robotic haircutting operating under experimentally measured communication conditions. The proposed system employs a smartphone-based master interface with ArUco marker tracking and a remote robotic manipulator connected via a cloud communication framework. Three reference reconstruction techniques, namely ZOH, FOH, and LPF, are experimentally evaluated together with a classical proportional-derivative controller and a passivity-based synchronization controller using real communication data. The main contributions of this paper are fourfold. First, we develop a cloud-based unilateral teleoperation framework for robotic haircutting that integrates smartphone vision-based target tracking with remote robotic manipulation, enabling practical cloud-assisted teleoperation over the Internet. Second, we experimentally characterize real Internet communication uncertainties by measuring key network performance metrics, including communication delay, jitter, packet timing variability, and maximum delay, thereby providing a realistic evaluation environment beyond simulated or idealized network models. Third, we systematically investigate the impact of control and signal reconstruction strategies by comparing conventional proportional–derivative (PD) control with passivity-based synchronization control and evaluating ZOH, FOH, and LPF methods for handling asynchronous packet reception under realistic communication conditions. Finally, we conduct comprehensive real-world teleoperation experiments to quantitatively evaluate tracking accuracy, synchronization quality, and control effort, demonstrating the effectiveness of the proposed framework and providing practical insights into robust cloud-based teleoperation for robotic haircutting.
	
	The remainder of this paper is organized as follows. Section II introduces the proposed teleoperation framework. Section III presents the communication characterization and delay-compensation methods. Section IV describes the control strategies. Section V reports the experimental validation and comparative results. Finally, Section VI concludes the paper.
	
	\section{System Architecture}
	
\begin{figure}
		\begin{center}
			\includegraphics[width=3.45in]{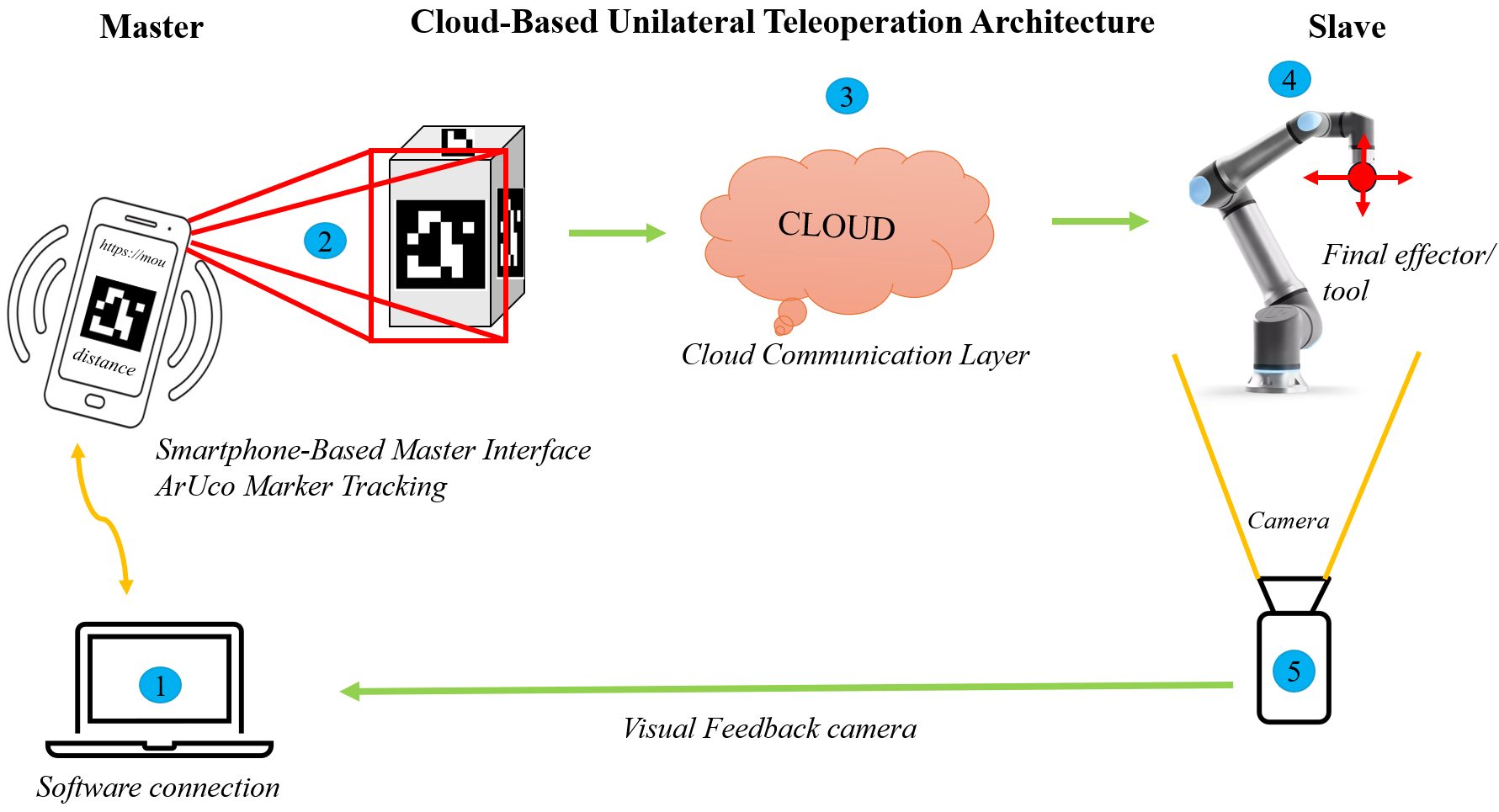}\\
			\caption{Proposed cloud-based unilateral teleoperation architecture. The system consists of (1) a local computer running the teleoperation software, (2) a smartphone-based master interface that estimates the pose of an ArUco marker in real time, (3) a cloud communication layer responsible for transmitting commands through the Internet, (4) a robotic manipulator acting as the slave system, and (5) a remote camera providing visual feedback to the operator. Communication uncertainties such as transmission delays, jitter, and asynchronous packet arrivals are introduced by the Internet communication channel.}\label{unilateral}
		\end{center}
	\end{figure}

	The proposed cloud-based unilateral teleoperation architecture is illustrated in Fig. \ref{unilateral}. The operator manipulates an ArUco marker observed by the smartphone camera, which continuously estimates the marker pose and transmits position and orientation information through the cloud communication framework. At the remote site, the received data are converted into motion commands for the robotic manipulator. A secondary camera provides visual feedback to the operator, enabling remote monitoring of the teleoperation task.
	
	Since the communication channel relies on public Internet infrastructure, the transmitted data are affected by time-varying delays, packet timing variability, and asynchronous packet arrivals. These communication effects constitute the primary source of uncertainty analyzed in this work.
	
	The smartphone-based master interface performs real-time pose estimation of an ArUco marker using the embedded camera. The estimated pose is timestamped and transmitted through a cloud communication framework to the remote robotic manipulator. The estimated master pose, denoted by $r_m$, is represented by
	
	\begin{equation}
		r_m=
		\begin{bmatrix}
			x_m & y_m & z_m &
			r_{x,m} & r_{y,m} & r_{z,m}
		\end{bmatrix}^{T},
	\end{equation}
	
	where $(x_m,y_m,z_m)$ denote the translational coordinates of the marker, and $(r_{x,m},r_{y,m},r_{z,m})$ denote the components of the rotation vector estimated by the ArUco pose estimation algorithm.
	
	The estimated master pose is transmitted through the cloud communication network and used as the motion reference for the remote robotic manipulator. The corresponding slave pose, denoted by $r_s$, is defined as
	
	\begin{equation}
		r_s=
		\begin{bmatrix}
			x_s & y_s & z_s &
			r_{x,s} & r_{y,s} & r_{z,s}
		\end{bmatrix}^{T},
	\end{equation}
	
	where $(x_s,y_s,z_s)$ denote the Cartesian position of the robot end-effector, and
	$(r_{x,s},r_{y,s},r_{z,s})$ denote the corresponding rotation vector returned by the ArUco pose estimation algorithm.

	The objective of the teleoperation system is to synchronize the slave pose with the master reference despite communication delays and network-induced uncertainties, such that
	
	\begin{equation}
		\lim_{t\rightarrow\infty}
		\left(r_m(t)-r_s(t)\right)=0.
	\end{equation}
	
	\begin{figure}
		\centering
		\includegraphics[width=3.7in]{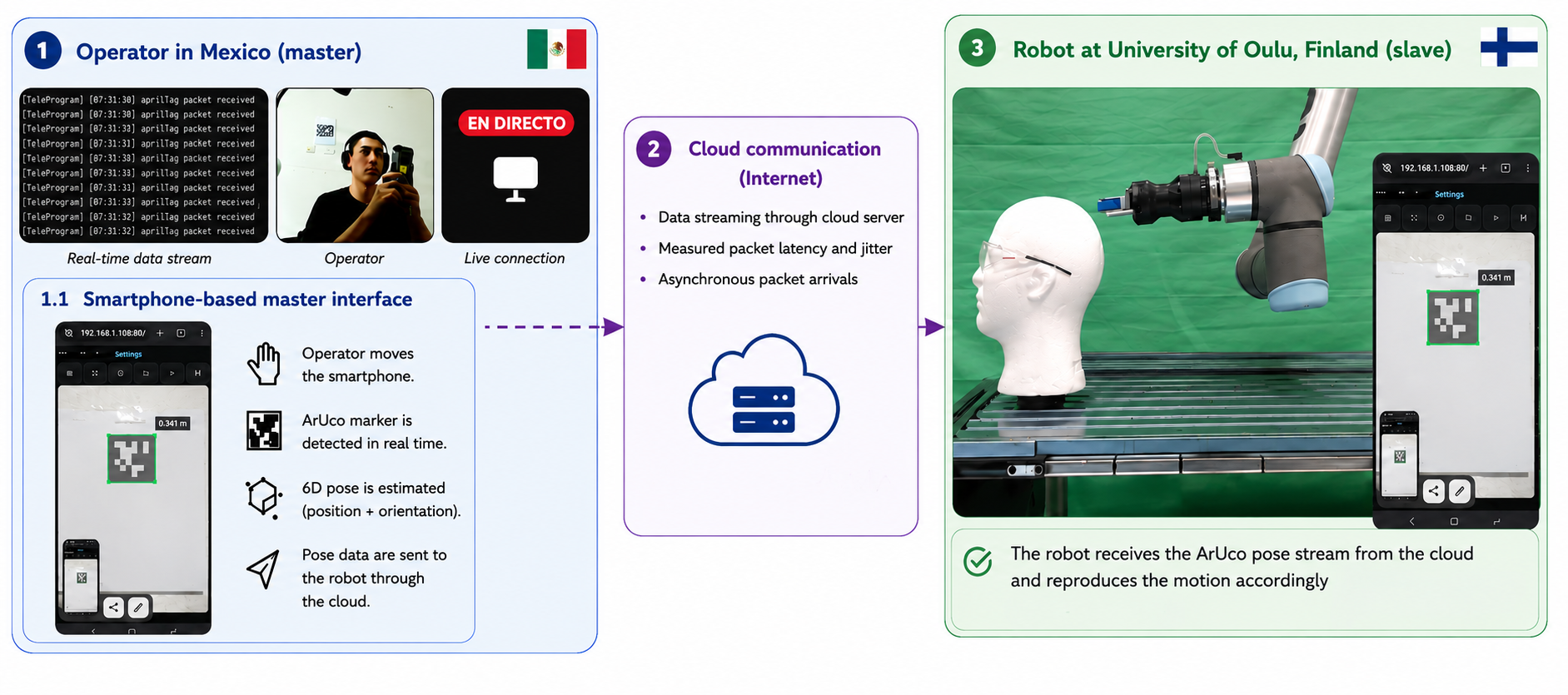}
		\caption{Experimental implementation of the cloud-based unilateral teleoperation platform.}
		\label{fig:experimentmxfin}
	\end{figure}
	
	The experimental implementation is shown in Fig. \ref{fig:experimentmxfin} of the proposed teleoperation platform used to collect the communication dataset employed throughout this work. The operator, located in Mexico, manipulated a smartphone-based master interface that estimated the pose of an ArUco marker in real time. The estimated pose was transmitted through a cloud communication framework to a UR robotic manipulator located at the University of Oulu, Finland, where the robot reproduced the received motion. A mannequin head was employed during these preliminary experiments to evaluate the teleoperation framework before experiments involving human participants safely. A detailed description of the complete teleoperation platform, communication framework, and software implementation can be found in \cite{Zhendai2026}.
	
	The complete implementation of the proposed teleoperation platform, including the cloud communication framework, smartphone-based master interface, and software architecture, is described in detail in \cite{Zhendai2026}. Fig. \ref{fig:experimentmxfin} presents the experimental implementation employed in this work to acquire the communication dataset used in the network characterization, delay reconstruction, and controller evaluation presented in the following sections. During the experiments, the operator was located in Mexico, while the slave robotic manipulator was deployed at the University of Oulu, Finland. The operator manipulated an ArUco marker using the smartphone-based master interface, whose estimated 6D pose was transmitted through the cloud communication framework and directly reproduced by the remote robot. A mannequin head was used as a preliminary validation platform to evaluate the teleoperation system before experiments involving human participants were conducted safely. All transmitted data, including timestamps and pose information, were recorded during these international teleoperation experiments and subsequently used for offline characterization of Internet communication uncertainties and evaluation of the proposed delay-compensation and control strategies.

	\subsection{Network Characterization}
	
	Unlike conventional teleoperation studies that assume constant communication delays, the proposed cloud-based teleoperation system operates over a real Internet connection, where transmission delays are inherently time-varying, and packets may arrive asynchronously.
	
	The transmission latency of packet $(p)$ can be expressed as
	
	\begin{equation}
		T_p(t)=\bar{T}+\Delta T_p(t),
	\end{equation}
	where $\bar{T}$ denotes the average packet transmission delay and $\Delta T_p(t)$ represents the time-varying deviation of packet $p$ from the average delay caused by network jitter and communication uncertainties.

\begin{figure}
		\centering
		\includegraphics[width=3.7in]{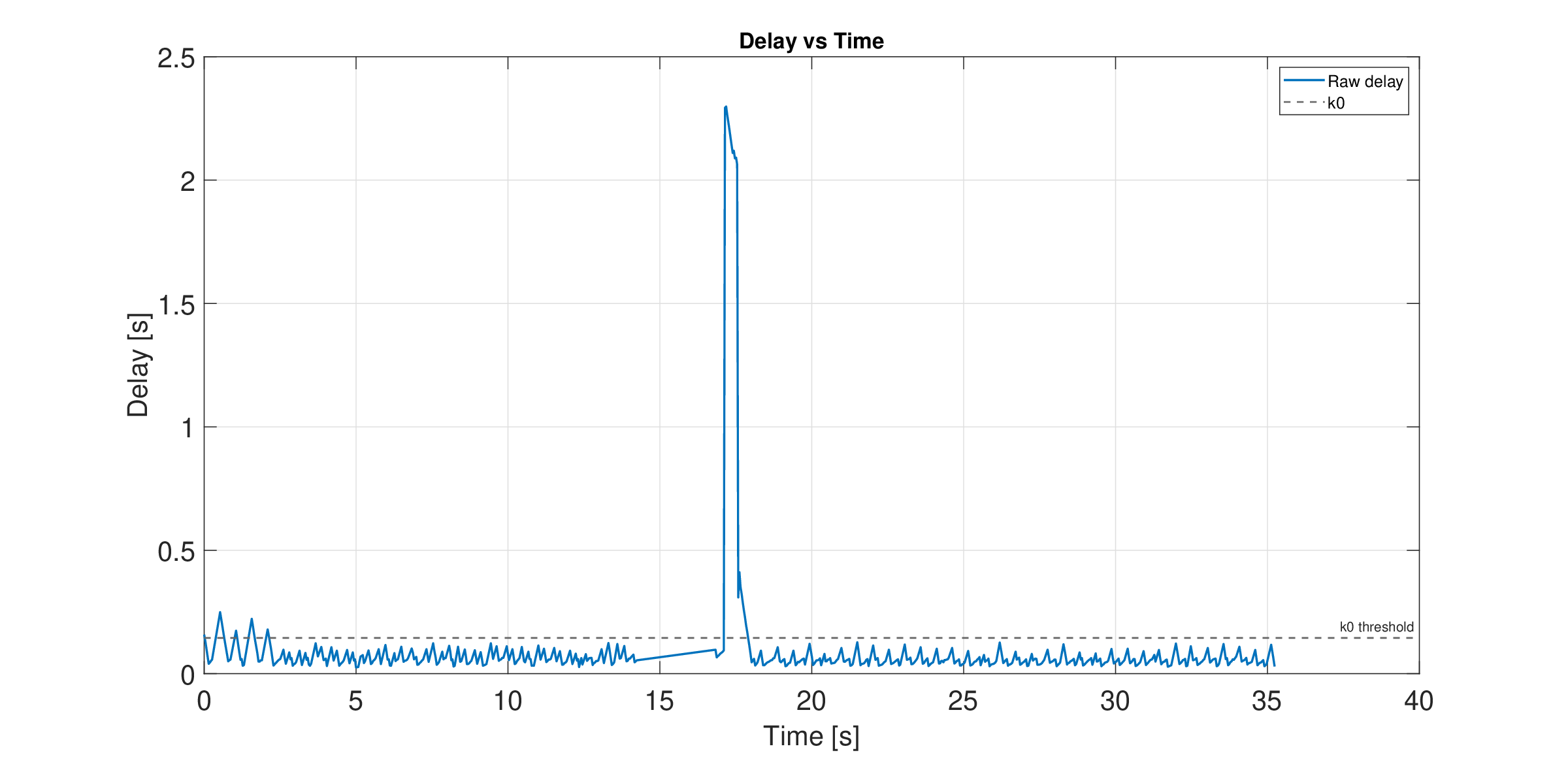}
		\caption{Measured packet transmission latency during the teleoperation experiment. The dashed line indicates the threshold $k_0 = 150$ ms used to identify late packets.}
		\label{fig:packet_delay}
	\end{figure}

	To characterize the communication channel, timestamps recorded at the smartphone client and the slave controller were used to estimate the end-to-end packet transmission latency during the teleoperation experiment. Fig.~\ref{fig:packet_delay} presents the measured packet latency over time.
	
	The packet latency remained below 150 ms for most of the experiment. However, an isolated communication disruption occurred around $t\approx17$~s, where several packets experienced transmission latencies exceeding 2~s. These values correspond to the individual travel time of delayed packets rather than the delay experienced by the controller itself. Packets whose transmission latency exceeded the threshold $k_0$ were classified as late arrivals and excluded from the reference reconstruction process.
	
\begin{figure}
		\centering
		\includegraphics[width=3.7in]{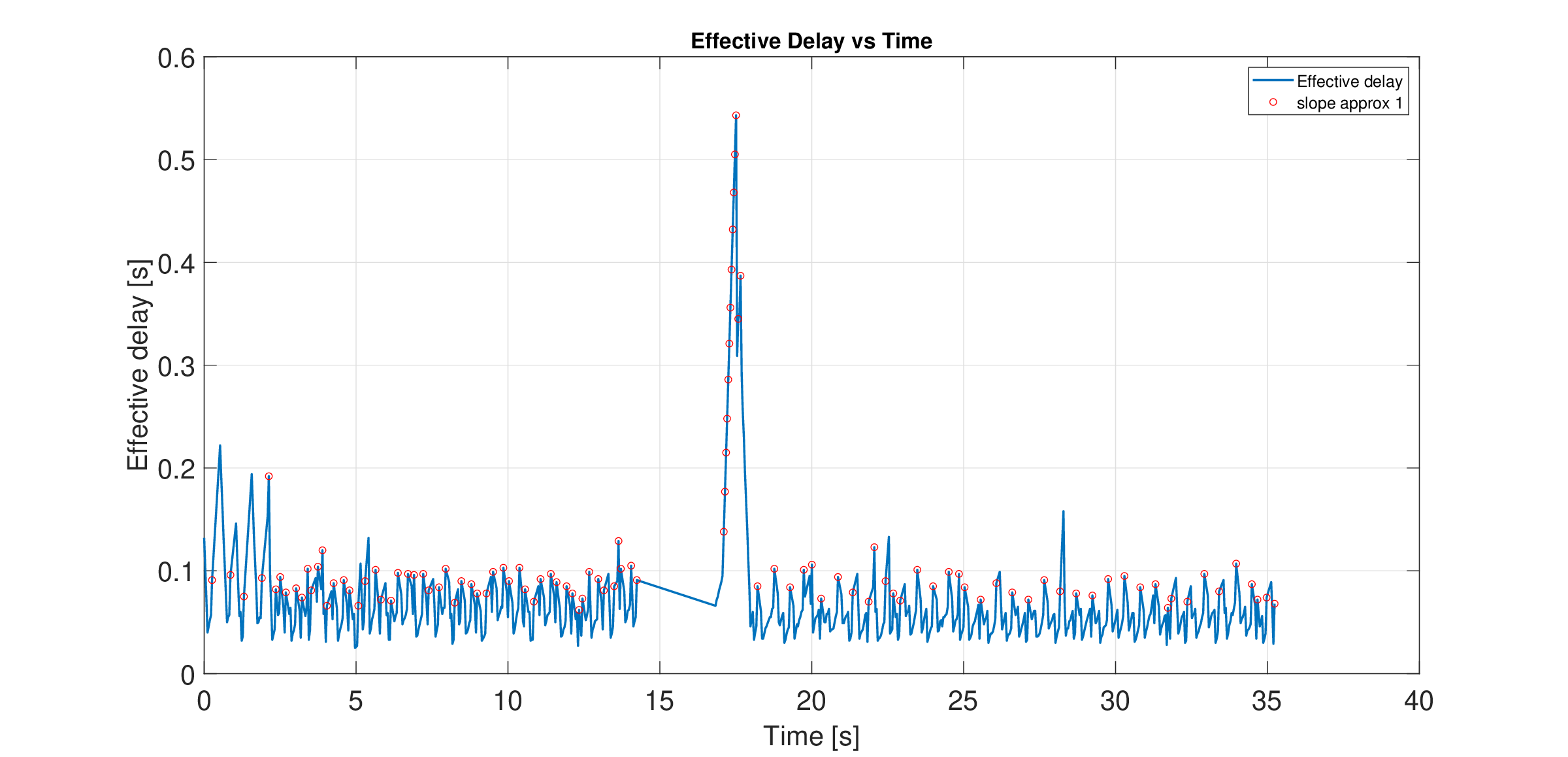}
		\caption{Effective communication delay perceived by the controller. Red markers indicate samples where the delay growth approaches the theoretical upper bound.}
		\label{fig:effective_delay}
	\end{figure}

	Since the controller continuously operates using the most recently received packet, the effective communication delay perceived by the control loop differs from the transmission latency of individual packets. The effective delay is defined as
	
	\begin{equation}
		T_e(t)=t-t_{\mathrm{send}}^{\mathrm{latest}},
	\end{equation}
	where $t_{\mathrm{send}}^{\mathrm{latest}}$ denotes the transmission time of the newest packet already available at the controller. Fig.~\ref{fig:effective_delay} presents the resulting effective delay.
	
	Unlike packet transmission latency, the effective delay remains bounded throughout the experiment because the controller always exploits the most recent available information. Even during the communication disruption around $t\approx17$~s, the effective delay increases only temporarily before rapidly recovering as new packets arrive.
	
	\begin{figure}
		\centering
		\includegraphics[width=3.7in]{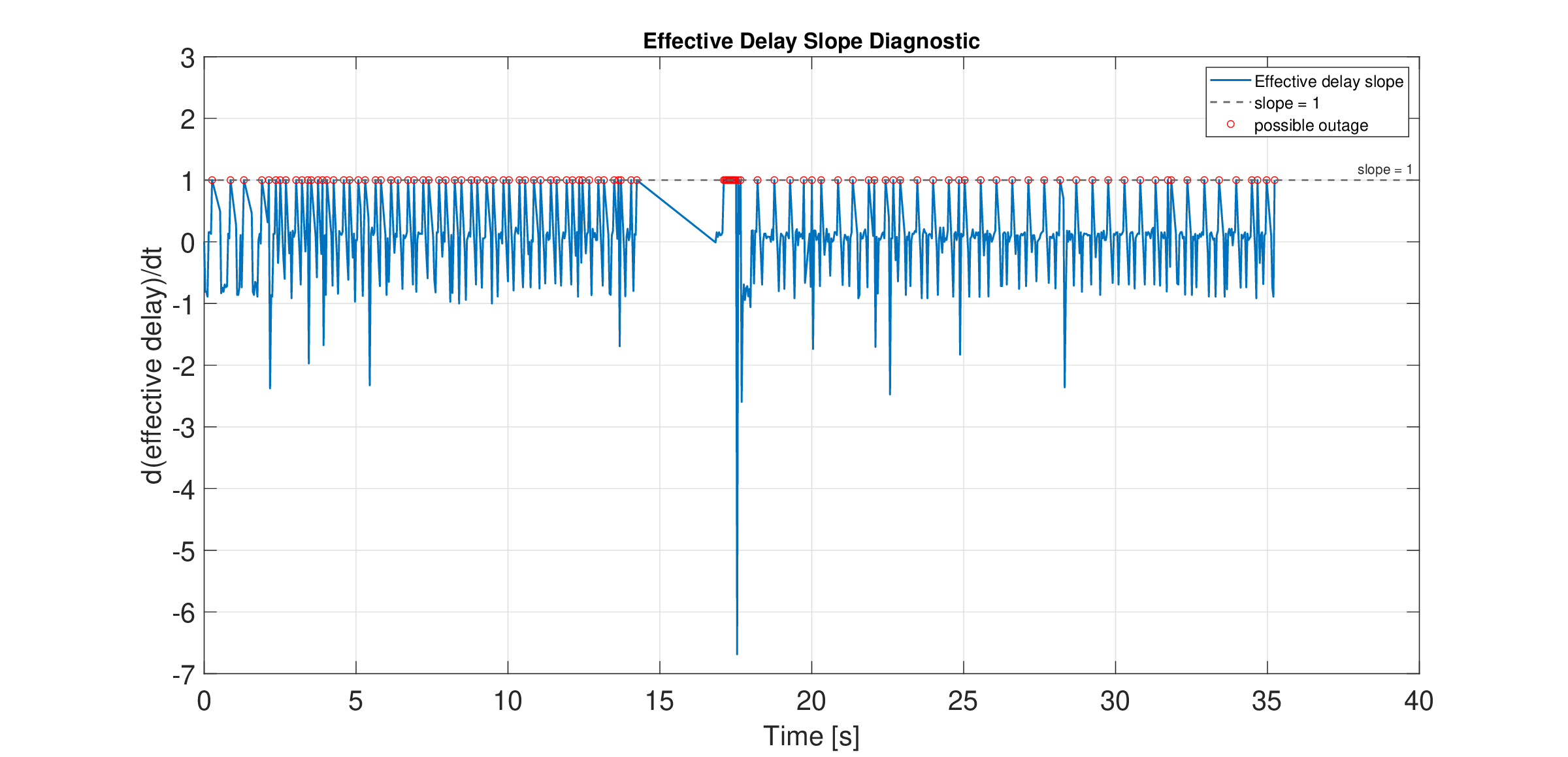}
		\caption{Derivative of the effective delay. Samples approaching a unit slope indicate intervals in which no new information becomes available and the age of the information increases at the same rate as time.}
		\label{fig:effective_delay_slope}
	\end{figure}
	
	To further verify the communication behavior, the derivative of the effective delay was evaluated. As shown in Fig.~\ref{fig:effective_delay_slope}, the slope of the effective delay remains below or close to the theoretical upper bound of one during the experiment. This behavior confirms that, although some packets experience long transmission delays, the controller continuously updates its state using the newest available measurements and therefore does not accumulate arbitrarily increasing communication delays.
	
	These observations demonstrate that the communication channel cannot be accurately represented by a constant-delay model. Instead, the experimental Internet connection exhibits variable packet latency, asynchronous packet arrivals, and temporary communication disruptions, motivating the use of delay compensation and robust teleoperation control strategies.
	
	\begin{table}[t]
		\caption{Experimental characterization of the Internet communication channel.}
		\centering
		\begin{tabular}{lc}
			\hline
			Metric & Value \\
			\hline
			Average packet delay & 100.9 ms \\
			Packet delay std. dev. & 279.9 ms \\
			Minimum packet delay & 25 ms \\
			Median packet delay & 55 ms \\
			95th percentile packet delay & 135 ms \\
			Maximum packet delay & 2.298 s \\
			Maximum effective delay & 543 ms \\
			Average sampling frequency & 19.41 Hz \\
			Late packets ($T_p>k_0$) & 4.82\% \\
			\hline
		\end{tabular}
		\label{tab:network_metrics}
	\end{table}
	
	Table~\ref{tab:network_metrics} summarizes the main communication characteristics observed during the experiments. Although the average packet transmission latency was approximately 100.9 ms, most packets experienced substantially lower delays, as reflected in the median of 55 ms. Sporadic communication disruptions produced packet transmission latencies of up to 2.298~s; however, these isolated events did not directly translate into equivalent control delays because the controller continuously operated using the most recently available packet. Consequently, the effective communication delay remained bounded throughout the experiment. Approximately 4.82\% of the packets exceeded the threshold $k_0$, confirming the presence of communication jitter and asynchronous packet arrivals typical of Internet-based teleoperation.

	
	\section {Delay reconstruction strategies}
	
	The observed communication uncertainties motivate the use of signal reconstruction mechanisms that mitigate the effects of asynchronous packet arrivals and delay variability. In this work, three commonly used reconstruction approaches are deployed: 
	ZOH, FOH, and LPF. Each reconstructed signal is subsequently used as the reference input for both the classical PD controller and the passive synchronization controller, enabling a direct comparison of their performance under identical communication conditions.
	
	\subsection{Zero-Order Hold (ZOH)}
	
	ZOH method is one of the most widely used signal reconstruction techniques in networked control systems due to its simplicity and low computational cost. Under this approach, the most recently received packet is maintained constant until a new packet becomes available.
	
	Let $r_k$ denote the reference pose received at time $t_k$. The reconstructed ZOH reference is defined as
	
	\begin{equation}
		r_{\mathrm{ZOH}}(t)=r_k,
		\qquad
		t_k \leq t < t_{k+1}.
	\end{equation}
	
	Therefore, the reference remains unchanged between consecutive packet arrivals. This strategy guarantees the continuous availability of a valid reference signal even when packets arrive asynchronously or when temporary communication interruptions occur.
	
	In the proposed teleoperation architecture, the ZOH method is applied to the received master pose before being provided to the remote robot controller. Consequently, the slave robot always tracks the latest available reference transmitted through the communication channel.
	
	Fig.~\ref{fig:ref_zoh_foh_lpf} illustrates the reconstructed references obtained using ZOH together with the raw received signal and the other reconstruction approaches. As expected, the ZOH strategy provides a robust mechanism for handling packet losses and irregular packet arrivals. However, the discontinuous nature of the reconstructed signal may cause sudden reference changes, increasing the control effort required by the teleoperation controller. For this reason, ZOH is used in this work as a baseline reconstruction strategy against which the FOH and LPF approaches are compared.

	\subsection{First-Order Hold (FOH)}
	
	FOH method extends the Zero-Order Hold (ZOH) approach by approximating the reference signal as a piecewise linear function between consecutive samples. Unlike ZOH, which maintains the most recently received sample constant until the next update, FOH augments the ZOH output with a linear correction term computed from the two most recent samples, thereby offering a direct comparison between the two reconstruction methods.
	
	Following the classical definition of digital control systems, let $r_k$ and $r_{k-1}$ denote two consecutive samples acquired at the periodic sampling instants $t_k$ and $t_{k-1}$, respectively. The FOH reconstruction is defined as
	
	\begin{equation}
		r_{\mathrm{FOH}}(t)
		=
		r_k
		+
		(r_k-r_{k-1})
		\frac{t-t_k}{T_s},
		\qquad
		0 \leq t-t_k < T_s,
	\end{equation}
	where $T_s$ denotes the nominal sampling period.
	
	This expression reconstructs the sampled signal by adding a linear correction term to the ZOH output, producing a piecewise-linear trajectory over each sampling interval. Consequently, the reconstructed signal evolves continuously between consecutive samples rather than exhibiting the staircase behaviour characteristic of ZOH, providing clearer contrast between the two methods.
	
	The magnitude of the FOH correction term is directly influenced by the selected sampling period $T_s$. As $T_s$ increases, the correction term becomes progressively smaller, and the reconstructed signal approaches the Zero-Order Hold behaviour. Conversely, smaller values of $T_s$ increase the contribution of the correction term, producing steeper linear extrapolations that may amplify transient deviations during communication interruptions, in contrast to the smoother ZOH response.
	
	In the proposed teleoperation framework, the received master pose is reconstructed using a causal FOH strategy before being supplied to the slave controller. Since the communication channel introduces irregular packet arrivals, the received reference is first represented as a ZOH signal, after which the FOH correction term is applied according to (6). During communication outages, this reconstruction continues to rely solely on the two most recent packets, making it suitable for real-time implementation without requiring future information.
	
	Fig.~\ref{fig:ref_zoh_foh_lpf} compares the reconstructed references obtained using FOH with the raw measurements, ZOH reconstruction, and LPF reconstruction. Under nominal communication conditions, the FOH reconstruction replaces the piecewise-constant behaviour of ZOH with piecewise-linear segments, resulting in smoother transitions between consecutive received samples. However, during prolonged communication interruptions, the elapsed time since the last received packet increases because no new measurements are available. Consequently, the FOH reconstruction behaves as a local linear extrapolation during the outage, which may temporarily deviate from the actual operator motion until a new packet is received.
	
	It should be noted that the classical FOH formulation assumes periodic sampling, and the reconstruction is valid only within a single sampling interval ($0 \leq t-t_k < T_s$). In cloud-based teleoperation, packet arrivals are inherently asynchronous, and communication outages may occur, so these assumptions are not always satisfied. Nevertheless, FOH remains a computationally efficient and causal reconstruction strategy that provides a continuous reference signal while requiring only the two most recently received samples.
	
	The effectiveness of the FOH reconstruction is experimentally evaluated in the following sections by comparing tracking performance, synchronisation accuracy, and control effort against the ZOH and LPF approaches.
	
	\subsection{Low-Pass Filtering (LPF)}
	
	While ZOH and FOH focus on reconstructing missing information between packet arrivals, LPF aims to attenuate the high-frequency fluctuations introduced by communication jitter and asynchronous packet reception. By smoothing abrupt variations in the received reference signal, LPF can reduce undesirable oscillations in the teleoperation system and generate smoother control actions.
	
	In this work, the received master reference is filtered using a first-order discrete low-pass filter defined as
	
	\begin{equation}
		r_{\mathrm{LPF}}[k]
		=
		\alpha\, r[k]
		+
		(1-\alpha)\, r_{\mathrm{LPF}}[k-1],
	\end{equation}
	where $r[k]$ denotes the received reference sample,
	$r_{\mathrm{LPF}}[k]$ is the filtered output,
	and $\alpha \in (0,1)$ is the filter coefficient
	that determines the trade-off between responsiveness and smoothing.
	
	The filtering operation attenuates high-frequency components caused by network-induced disturbances while preserving the dominant motion of the operator. Consequently, the reconstructed reference becomes smoother than those obtained using ZOH and FOH, particularly during intervals characterized by irregular packet arrivals.
	
\begin{figure}[!t]
		\centering
		\includegraphics[width=3.7in]{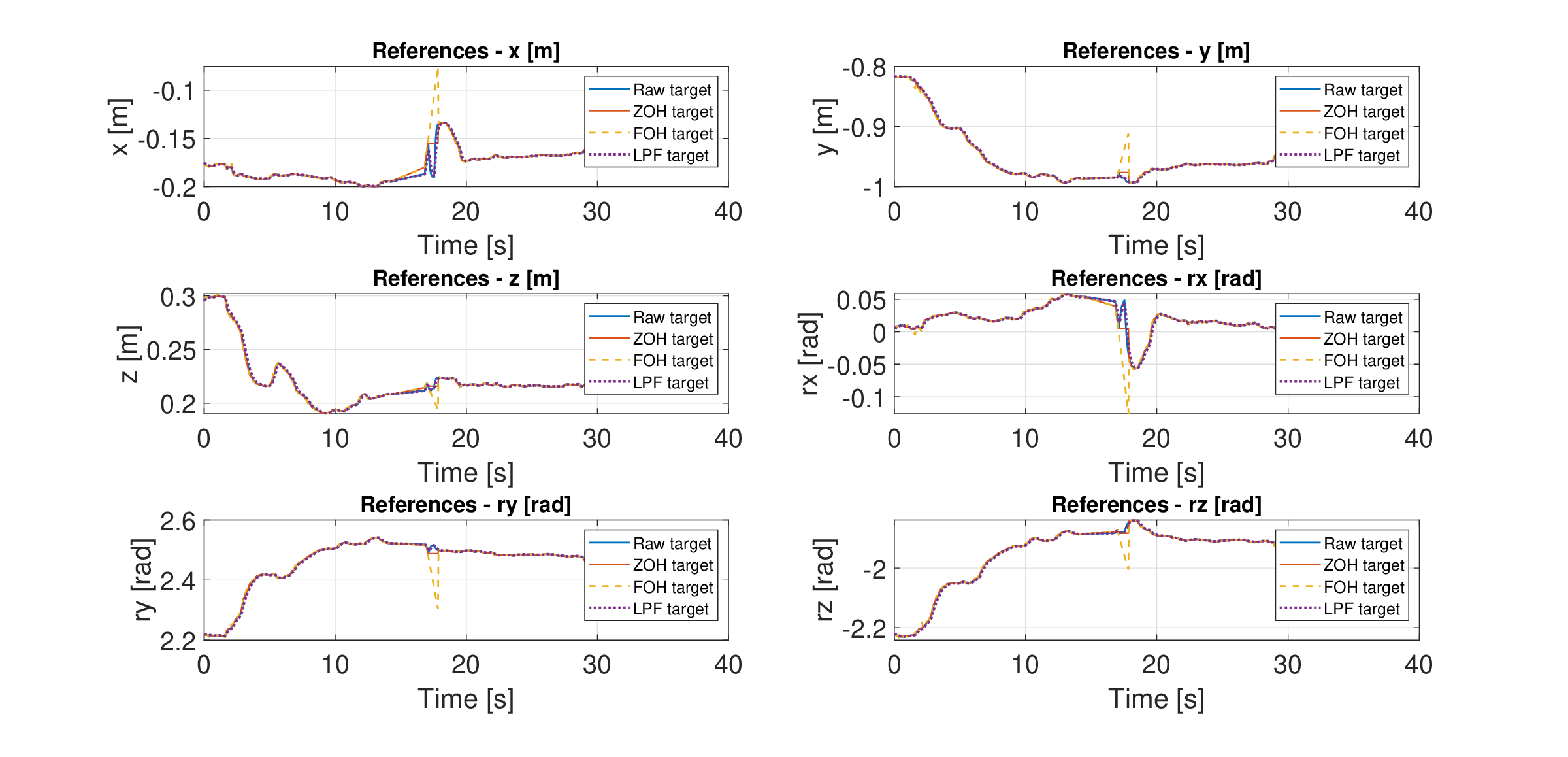}
		\caption{Comparison of the raw reference signal and the reconstructed references obtained using ZOH, FOH, and LPF around the largest communication disturbance observed during the experiment. ZOH preserves the last received sample and produces piecewise-constant segments. FOH reconstructs the reference using first-order extrapolation, which generates transient overshoots during the communication interruption. LPF attenuates communication-induced fluctuations while remaining close to the original reference trajectory, providing the smoothest reconstructed signal.}
		\label{fig:ref_zoh_foh_lpf}
	\end{figure}

	Fig.~\ref{fig:ref_zoh_foh_lpf} presents the LPF-reconstructed references together with the raw received signal and the other reconstruction methods. The LPF reconstruction attenuates abrupt variations and high-frequency fluctuations introduced by communication jitter while remaining close to the original reference trajectory. Unlike ZOH, which preserves the last received sample, and FOH, which generates linear extrapolations that may produce transient overshoots during prolonged communication interruptions, LPF provides the smoothest reference reconstruction among the three methods.
	
	The main advantage of LPF is its ability to generate smooth reference trajectories that reduce control effort while preserving tracking performance. Although low-pass filtering inherently introduces additional phase lag, the selected cutoff frequency in this work provides sufficient bandwidth for the considered teleoperation task. Consequently, no significant degradation of the tracking accuracy is observed, while communication-induced fluctuations are effectively attenuated.
	
	The influence of LPF on pose synchronization performance and control effort is experimentally investigated in the following sections and compared against the ZOH and FOH reconstruction strategies.

	\begin{table}[t]
		\caption{Characteristics of the signal reconstruction strategies.}
		\centering
		\begin{tabular}{lccc}
			\hline
			Property & ZOH & FOH & LPF \\
			\hline
			Implementation complexity & Low & Medium & Low \\
			Signal continuity & No & Yes & Yes \\
			Jitter attenuation & Low & Medium & High \\
			Transient overshoot & None & Possible & Low \\
			Additional lag & None & None & Small \\
			Robustness to communication interruptions & High & Medium & High \\
			\hline
		\end{tabular}
		\label{tab:reconstruction_comparison}
	\end{table}
	
	\section{Controllers}
	
	To evaluate the influence of communication uncertainties on teleoperation performance, two different control strategies are considered in this work. The first controller corresponds to a classical PD approach, which represents one of the most widely used solutions in practical teleoperation systems due to its simplicity and ease of implementation. The second controller is based on passive synchronization, a passivity-inspired framework that seeks to preserve synchronization between the remote systems while maintaining a dissipative closed-loop behavior under communication delays.
	
	The reconstructed reference described in the previous section is received by both controllers, consequently, the comparison focuses exclusively on the influence of the control strategy under identical experimental conditions. Unlike previous studies that evaluate passivity-based synchronization under idealized communication assumptions, both controllers are implemented using the same experimentally measured Internet delays and identical reference reconstruction strategies. This experimental design isolates the effect of the control algorithm from the effects introduced by the communication channel, enabling a fair comparison between the classical PD controller and the passivity-based synchronization approach.
	
	\subsection{Classical PD Controller}
	
	The classical PD controller computes the control action as a function of the position and velocity tracking errors between the received reference pose and the measured slave robot pose \cite{Abut2017}.
	
	Let
	
	\begin{equation}
		e(t)=r_m(t)-r_s(t)
	\end{equation}
	denote the pose tracking error, where $r_m(t)$ represents the reconstructed master reference and $r_s(t)$ corresponds to the measured slave pose.
	
	The control law is given by
	
	\begin{equation}
		u_{PD}(t)
		=
		K_p e(t)
		+
		K_d \left(
		\delta_k \dot r_d(t)
		-
		\dot r_s(t)
		\right)
		-
		D_l \dot r_s(t),
	\end{equation}
	where $K_p$ and $K_d$ are diagonal proportional and derivative gain matrices, respectively.
	
	The proportional term generates a corrective action proportional to the instantaneous tracking error, while the derivative term introduces damping and improves the transient response. Although the PD controller provides satisfactory performance under nominal communication conditions, its tracking performance may deteriorate when large communication delays and asynchronous packet arrivals are present.

	For this reason, the PD controller is used in this work as a baseline teleoperation strategy against which the passive synchronization approach is compared. The controller gains used throughout the experiments are summarized in Table~\ref{tab:pd_parameters}.
	
	\begin{table}[t]
		\caption{Parameters of the classical PD controller.}
		\centering
		\begin{tabular}{lc}
			\hline
			Parameter & Value\\
			\hline
			$K_p$ (position)      & 4 \\
			$K_p$ (orientation)   & 2 \\
			$K_d$ (position)      & 0.10 \\
			$K_d$ (orientation)   & 0.05 \\
			$D_l$ (position)      & 0.15 \\
			$D_l$ (orientation)   & 0.08 \\
			\hline
		\end{tabular}
		\label{tab:pd_parameters}
	\end{table}
	
	\subsection{Passive Synchronization Controller}
	
	To address communication uncertainties, a passive synchronization controller is implemented following the synchronization framework proposed by Chopra and Spong \cite{Chopra2005}. Instead of directly synchronizing positions, the controller defines passive synchronization variables that combine position and velocity information. 
	
	The master synchronization output is defined as
	
	\begin{equation}
		r_m(t)
		=
		\delta_k \dot r_d(t)
		+
		\Lambda r_d(t),
	\end{equation}
	where $r_d(t)$ denotes the reconstructed reference pose,
	$\Lambda$ is a positive-definite synchronization gain matrix,
	and $\delta_k$ is a packet-availability indicator that takes the value
	
	\begin{equation}
		\delta_k=
		\begin{cases}
			1, & \text{new packet available}\\
			0, & \text{otherwise}
		\end{cases}
	\end{equation}
	
	The slave synchronization output is defined as
	
	\begin{equation}
		r_s(t)
		=
		\dot r_s(t)
		+
		\Lambda r_s(t).
	\end{equation}
	
	The passive synchronization error is then computed as
	
	\begin{equation}
		e_r(t)
		=
		r_m(t)-r_s(t).
	\end{equation}
	
	The control law applied to the slave robot is given by
	
	\begin{equation}
		u_{PS}(t)
		=
		K_s e_r(t)
		-
		D_s \dot r_s(t),
	\end{equation}
	where $K_s$ is the synchronization gain matrix and $D_s$ is a positive-definite damping matrix.
	
	The synchronization error combines both position and velocity information, allowing the controller to preserve coordinated behavior despite communication delays and asynchronous packet arrivals. The damping term contributes to energy dissipation and improves the robustness of the synchronization process under communication uncertainties.

	Unlike the classical PD controller, the passive synchronization strategy explicitly enforces synchronization between passive outputs rather than directly minimizing the position error. This formulation provides a passivity-inspired synchronization mechanism that is well suited for teleoperation systems operating under communication uncertainties.
	
	The parameters adopted for the passive synchronization controller are listed in Table~\ref{tab:passive_parameters}. These gains were independently tuned to obtain stable synchronization under the measured Internet communication conditions while reducing the control effort.
	
	\begin{table}[t]
		\caption{Parameters of the passive synchronization controller.}
		\centering
		\begin{tabular}{lc}
			\hline
			Parameter & Value\\
			\hline
			$\Lambda$ (position)    & 25 \\
			$\Lambda$ (orientation) & 18 \\
			$K_s$ (position)        & 0.06 \\
			$K_s$ (orientation)     & 0.03 \\
			$D_s$ (position)        & 0.20 \\
			$D_s$ (orientation)     & 0.10 \\
			\hline
		\end{tabular}
		\label{tab:passive_parameters}
	\end{table}

	Although both controllers are evaluated under the same communication conditions, their performance will be compared in terms of tracking accuracy and control effort. The following experimental section analyzes these differences by combining each controller with the proposed delay-compensation strategies.
	
	
	\section {Experimental Results}
	The performance of both controllers is experimentally evaluated using the previously described cloud-based teleoperation platform. The analysis focuses on three aspects: tracking accuracy, synchronization quality, and control effort. For each controller, the reconstructed references generated by ZOH, FOH, and LPF are used under identical communication conditions, allowing a direct assessment of the influence of both the control strategy and the delay compensation mechanism.
	
	\subsection{Experimental Setup}
	
	The experimental platform consists of a smartphone-based master interface and a remote robotic manipulator, connected via a cloud-based communication framework. The operator manipulates an ArUco marker whose pose is estimated in real time by the smartphone camera. The estimated pose is transmitted through the Internet to the cloud server and subsequently delivered to the slave robot controller.
	
	All experiments were performed using the same recorded communication data, including the measured transmission delays, packet timing irregularities, and asynchronous packet arrivals described in Section III. The average communication delay was approximately 100.9 ms, while the maximum observed delay reached 2.298 s.
	
	The reconstructed reference signals generated by the ZOH, FOH, and LPF strategies were evaluated using both the classical PD controller and the passive synchronization controller. Performance was assessed using pose-tracking accuracy and control-effort metrics.
	
	\subsection{PD Controller Performance}
	
\begin{figure}
		\centering
		\includegraphics[width=3.45in]{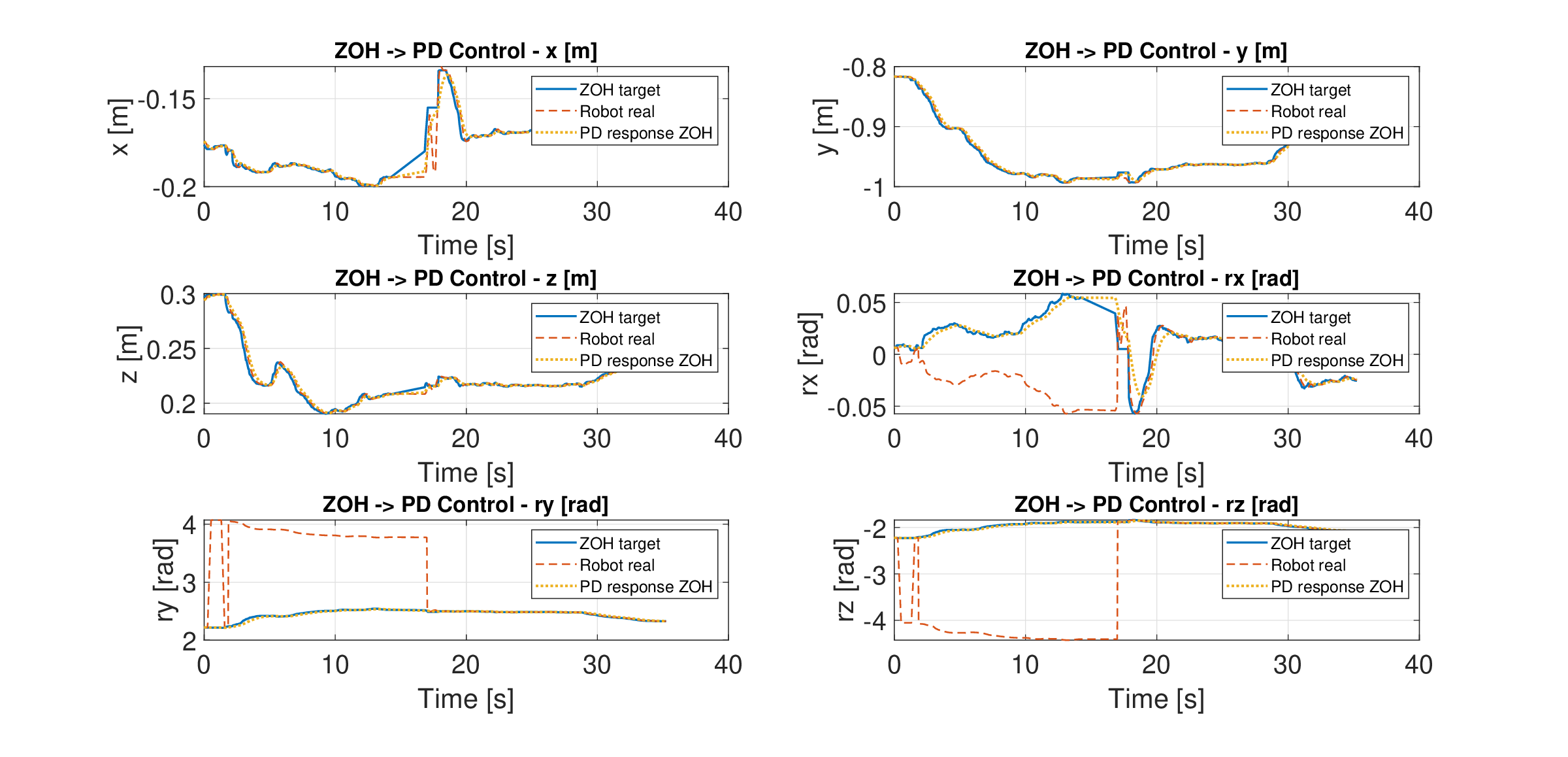}
		\caption{Tracking performance of the PD controller using the ZOH reconstruction strategy.}
		\label{fig:pd_zoh}
	\end{figure}

\begin{figure}
		\centering
		\includegraphics[width=3.45in]{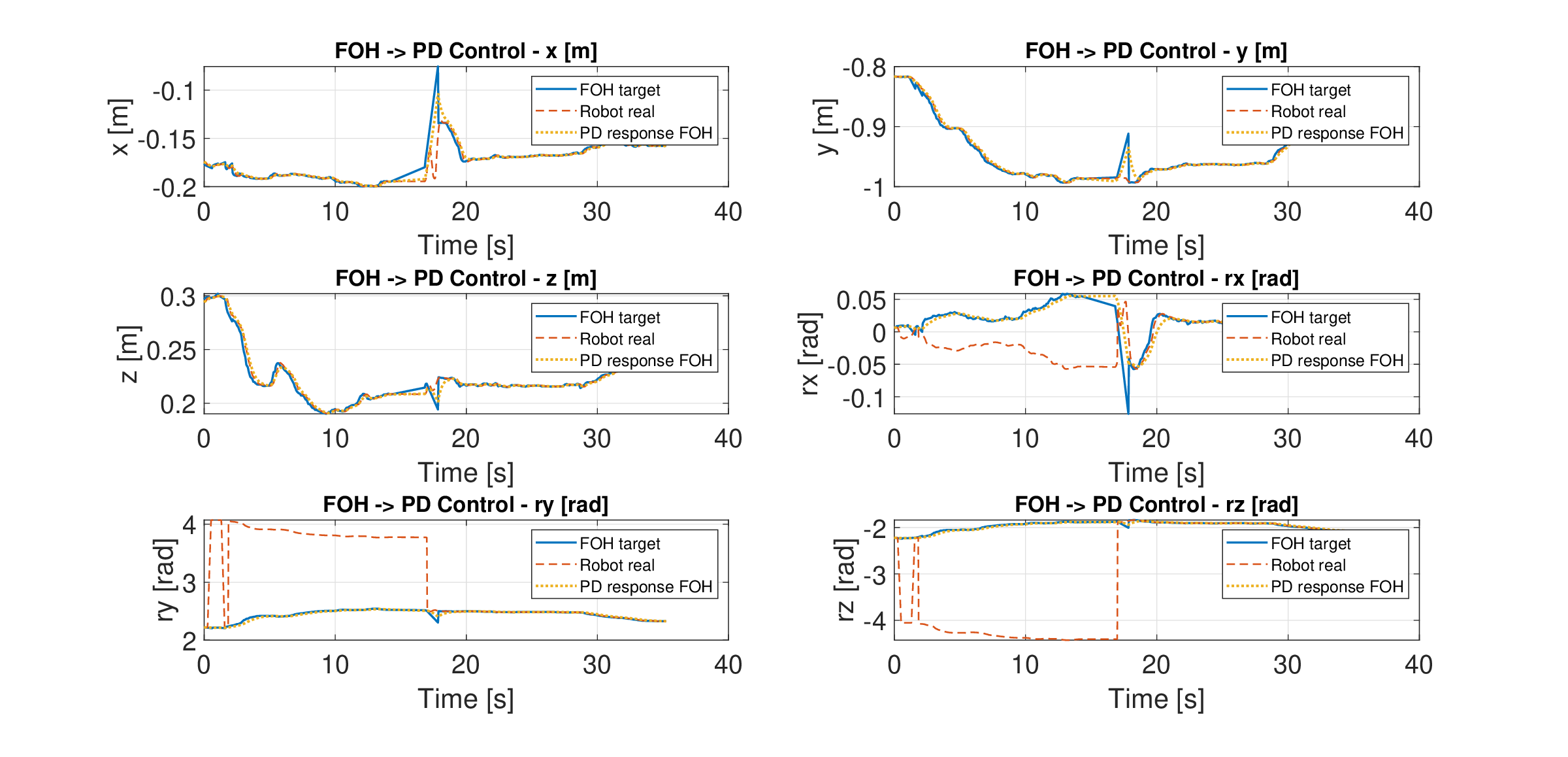}
		\caption{Tracking performance of the PD controller using the FOH reconstruction strategy.}
		\label{fig:pd_foh}
	\end{figure}

\begin{figure}
		\centering
		\includegraphics[width=3.45in]{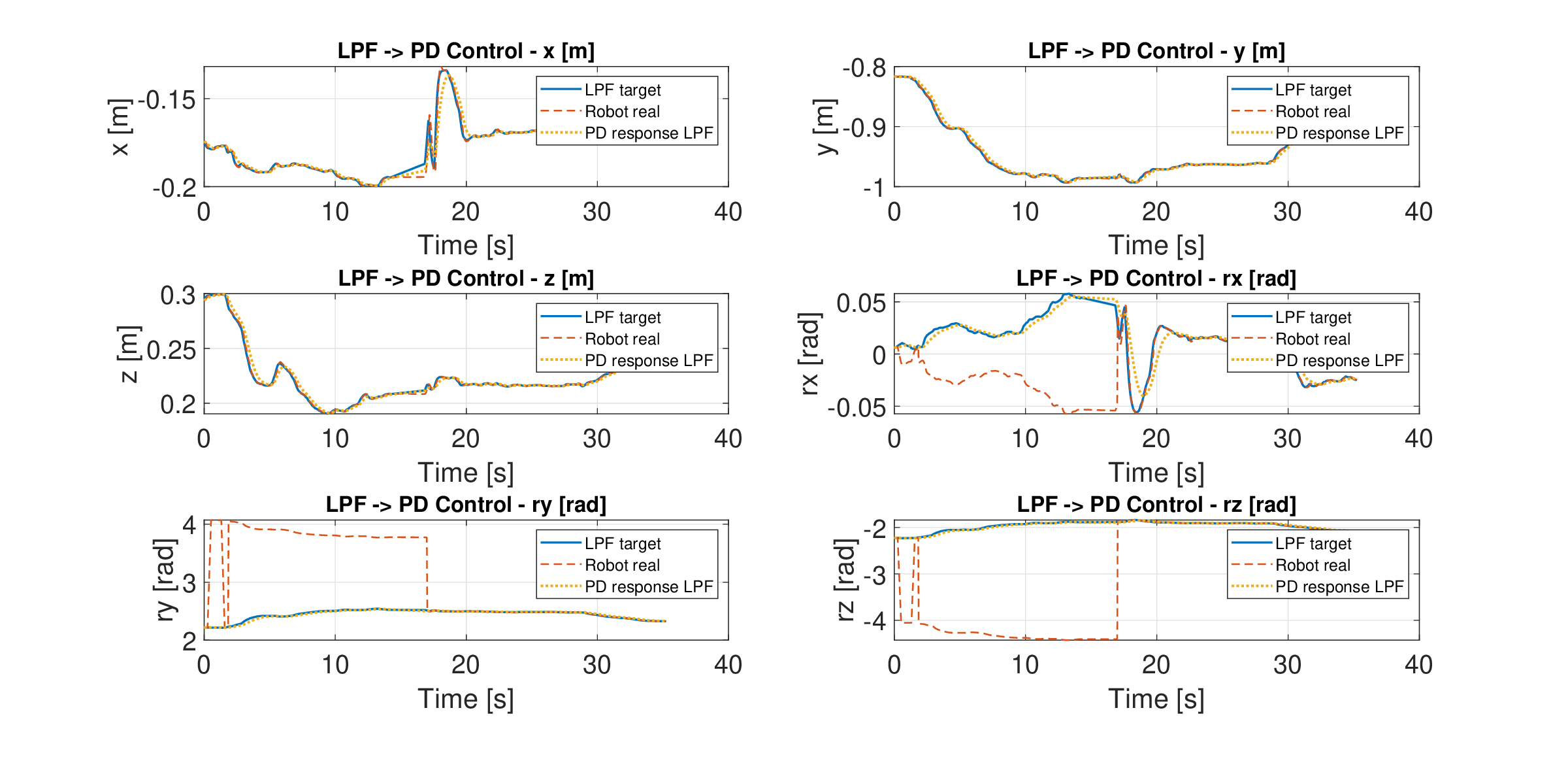}
		\caption{Tracking performance of the PD controller using the LPF reconstruction strategy.}
		\label{fig:pd_lpf}
	\end{figure}

	Figs.~\ref{fig:pd_zoh}--\ref{fig:pd_lpf} present the tracking performance obtained with the classical PD controller using the three reference reconstruction strategies.
	
	For all three reconstruction methods, the translational coordinates $(x,y,z)$ accurately follow the reconstructed references throughout most of the experiment. The largest tracking deviations occur around $t\approx17$~s, corresponding to the communication disturbance identified during the network characterization stage. Despite this temporary degradation, the controller rapidly recovers the desired trajectory and preserves stable operation.
	
	The influence of the reconstruction strategy becomes more evident during the communication disturbance. As expected, the ZOH reconstruction introduces abrupt reference changes due to its piecewise-constant behavior. FOH replaces these discontinuities with linear transitions; however, it also generates larger transient excursions during the delayed communication interval because of its linear extrapolation. In contrast, LPF provides the smoothest reference evolution by attenuating high-frequency variations while preserving the overall motion of the operator. 
	
	A similar behavior is observed for the rotational coordinates. The largest transient appears in the $r_{x,s}$ component, where FOH produces the most pronounced peak, whereas LPF effectively attenuates this oscillation. The $r_{y,s}$ and $r_{z,s}$ components remain comparatively smooth for all reconstruction strategies, with only minor deviations during the communication disturbance.
	
	The experimental robot measurements closely follow the simulated PD response for the translational coordinates, confirming that the controller reproduces the expected closed-loop behavior under real communication conditions. Larger discrepancies are observed in some rotational components, which could be influenced by unmodeled robot dynamics, orientation representation, or mechanical limitations rather than by the communication channel itself.
	
	Overall, all three reconstruction strategies maintain stable operation of the PD controller under realistic Internet communication conditions. Their quantitative differences in tracking accuracy and control effort are analyzed in the following subsection.
	
	\subsection{Passive Synchronization Controller Performance}
	
\begin{figure}
		\centering
		\includegraphics[width=3.45in]{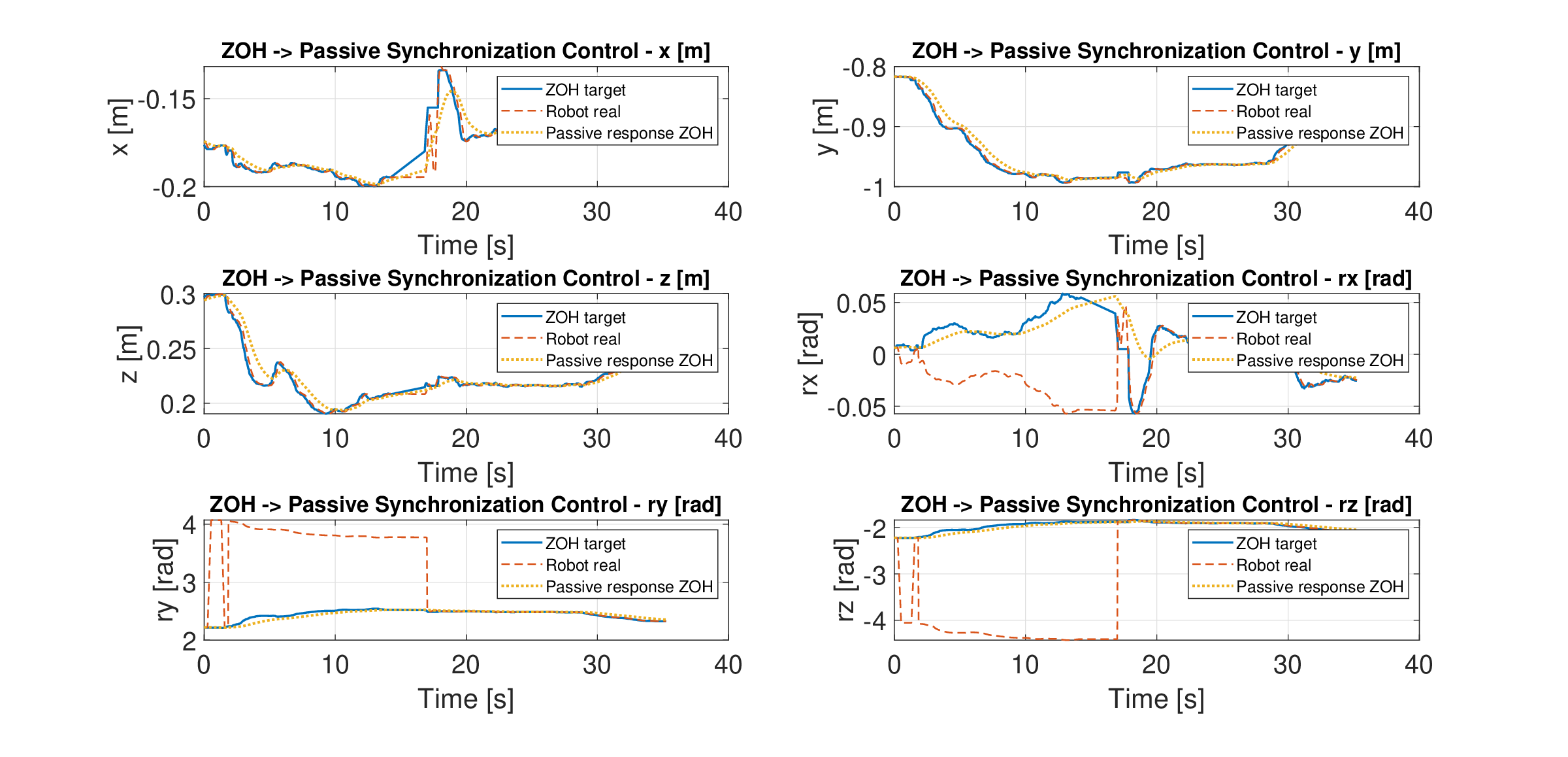}
		\caption{Tracking performance of the passive synchronization controller using the ZOH reconstruction strategy.}
		\label{fig:passive_zoh}
	\end{figure}

\begin{figure}
		\centering
		\includegraphics[width=3.45in]{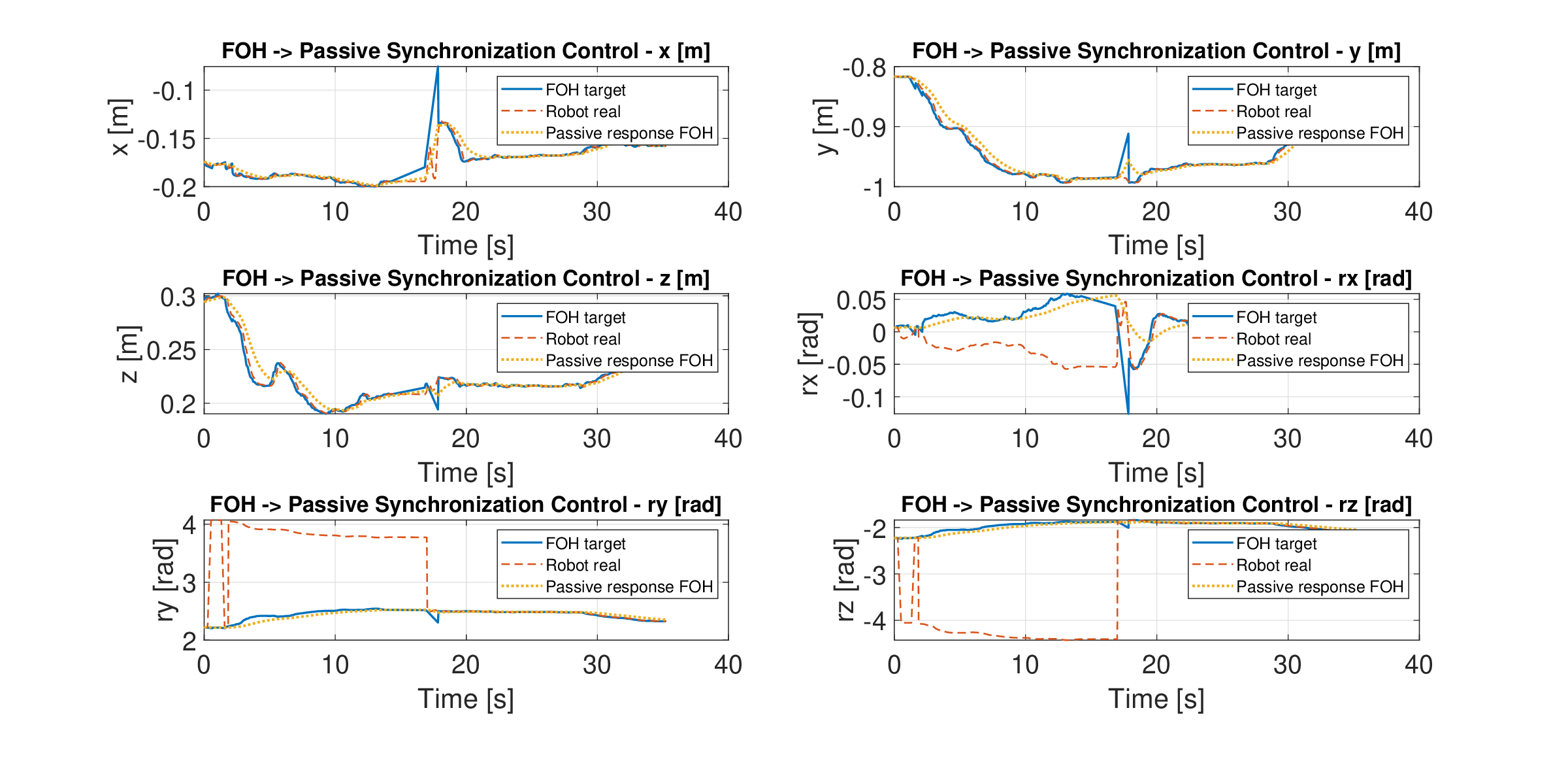}
		\caption{Tracking performance of the passive synchronization controller using the FOH reconstruction strategy.}
		\label{fig:passive_foh}
	\end{figure}

\begin{figure}
		\centering
		\includegraphics[width=3.45in]{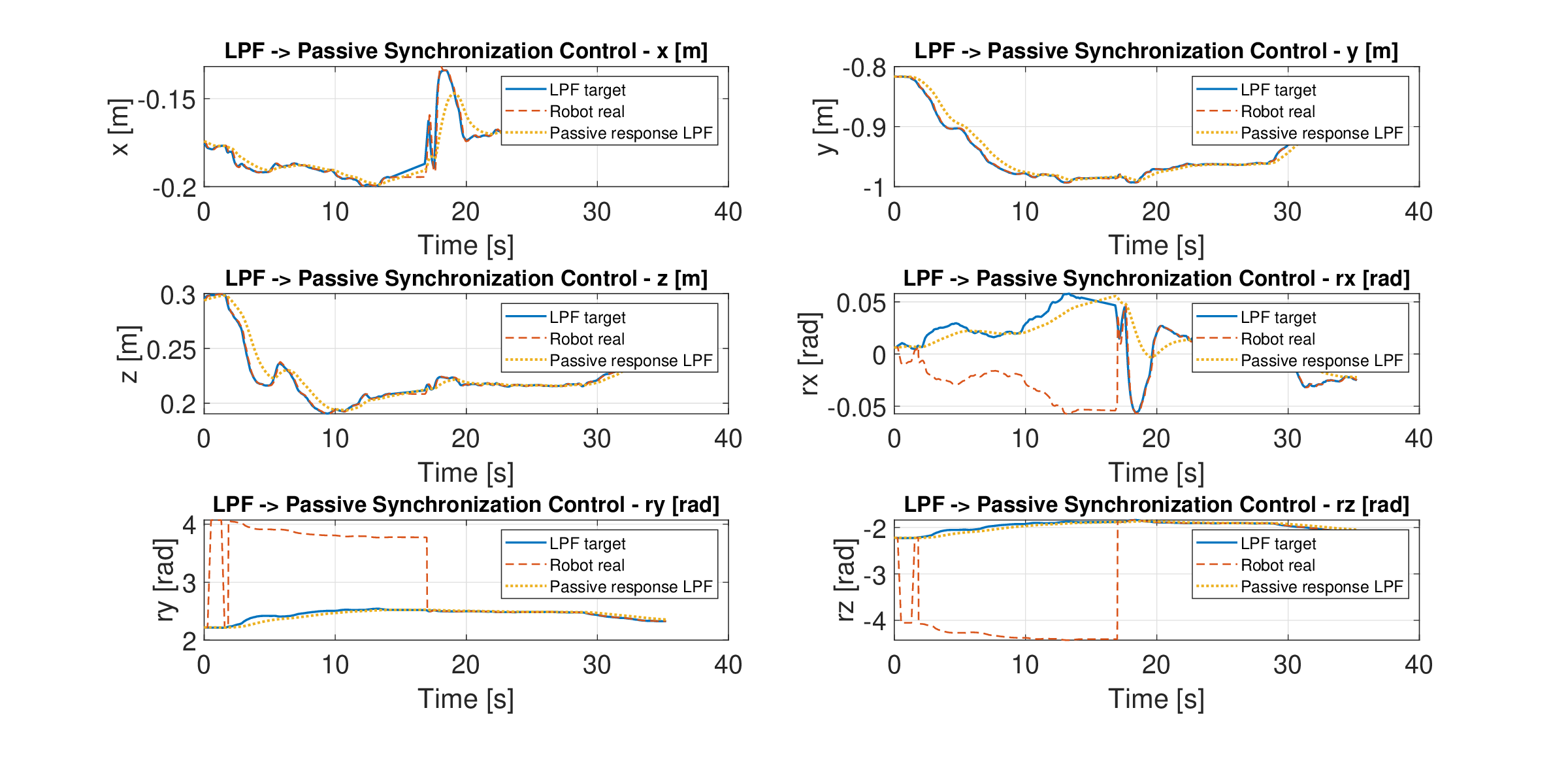}
		\caption{Tracking performance of the passive synchronization controller using the LPF reconstruction strategy.}
		\label{fig:passive_lpf}
	\end{figure}

	Figs.~\ref{fig:passive_zoh}--\ref{fig:passive_lpf} present the tracking performance obtained with the passive synchronization controller using the three reference reconstruction strategies.
	
	For all reconstruction strategies, the passive synchronization controller maintains stable closed-loop operation throughout the experiment. The translational coordinates $(x,y,z)$ accurately follow the reconstructed references under ZOH, FOH, and LPF, demonstrating that the synchronization framework remains effective under realistic Internet communication conditions.
	
	As observed for the classical PD controller, the largest tracking deviations occur around $t\approx17$~s, corresponding to the communication disturbance identified during the network characterization stage. Despite this temporary degradation, the synchronization controller rapidly converges back to the desired trajectory without compromising closed-loop stability.
	
	The influence of the reconstruction strategy is also consistent with the observations obtained for the PD controller. ZOH preserves the piecewise-constant behavior of the received reference, FOH replaces the discontinuities with linear transitions but produces the largest transient deviations during the communication disturbance, whereas LPF generates the smoothest reference evolution by attenuating high-frequency variations introduced by irregular packet arrivals.
	
	A similar behavior is observed for the rotational coordinates. The largest transient appears in the $rx$ component, while the $ry$ and $rz$ components remain comparatively smooth throughout the experiment. Although small deviations are visible during the communication disturbance, the synchronization controller maintains continuous rotational behavior for all reconstruction strategies.
	
	The experimental robot measurements closely follow the simulated passive synchronization response for the translational coordinates, confirming the validity of the proposed implementation under real communication conditions. As in the PD controller, larger discrepancies are mainly observed in the rotational variables because of the robot kinematics and orientation representation rather than the communication channel itself.
	
	Overall, the passive synchronization controller provides stable pose synchronization under realistic Internet communication conditions for all three reconstruction strategies. The quantitative comparison of tracking accuracy and control effort is presented in the following subsection.
	
	\subsubsection{Passive Synchronization Error}
	
	In addition to the pose tracking analysis, the synchronization error associated with the passive synchronization framework was evaluated. The synchronization error is defined as
	
	\begin{equation}
		e_r(t)=r_m(t)-r_s(t),
	\end{equation}
	where $r_m(t)$ and $r_s(t)$ denote the passive synchronization outputs defined in Section IV.
	
	\begin{figure}
		\centering
		\includegraphics[width=3.45in]{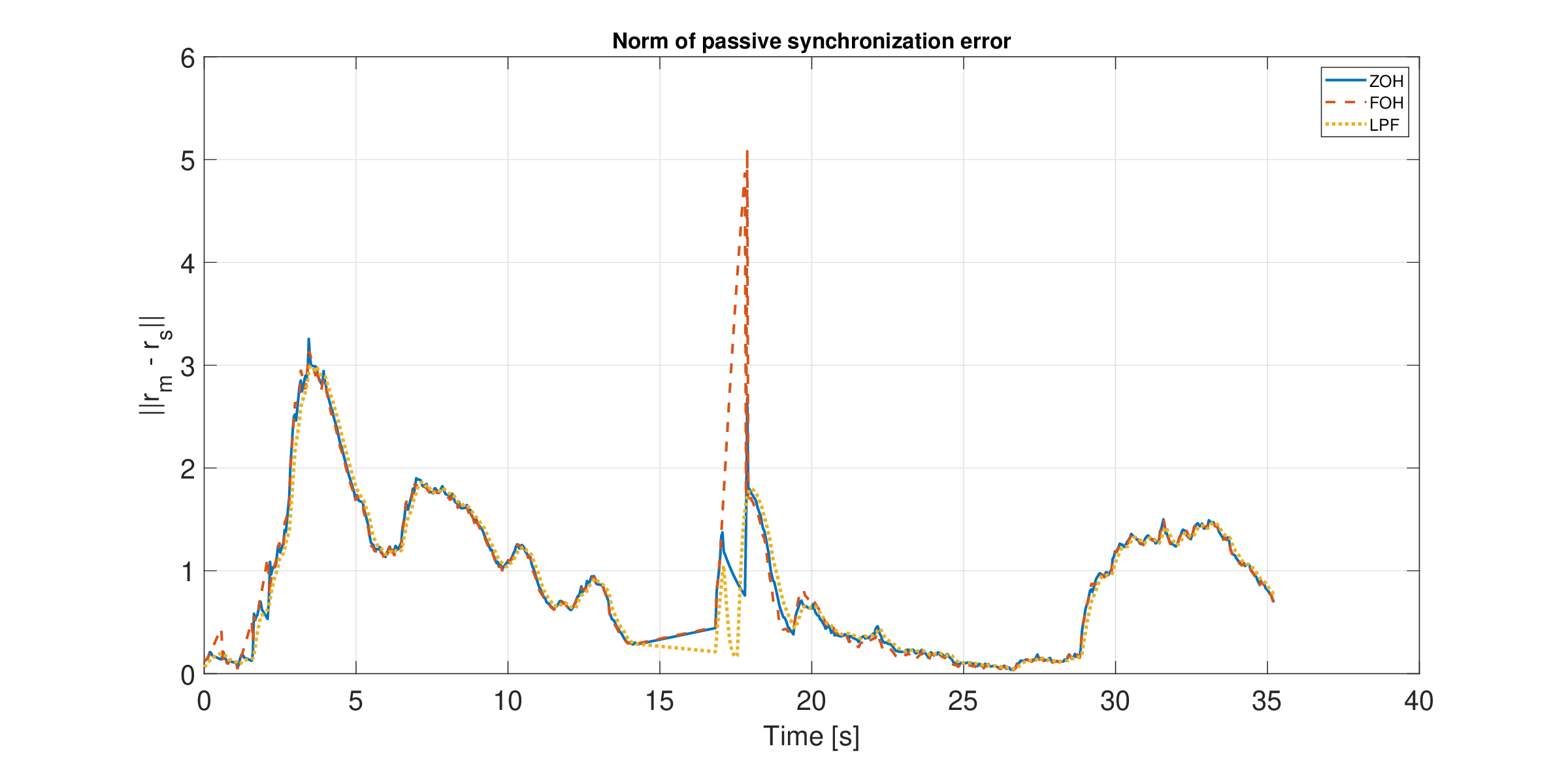}
		\caption{Norm of the passive synchronization error for the ZOH, FOH, and LPF reconstruction strategies.}
		\label{fig:norm_Passive_error}
	\end{figure}
	
	Fig.~\ref{fig:norm_Passive_error} shows the norm of the synchronization error for the three reconstruction strategies. The largest synchronization error occurs around the communication disturbance at $t\approx17$~s, coinciding with the interval where irregular packet arrivals temporarily degrade the quality of the reconstructed reference.
	
	Despite this communication disturbance, the synchronization error remains bounded throughout the experiment, confirming the robustness of the synchronization framework under realistic Internet communication conditions. Differences among the three reconstruction strategies are relatively small during nominal operation, while larger transient responses appear only during the communication disturbance. A quantitative comparison of the tracking performance is presented in Table~\ref{tab:tracking_rmse}.
	
	\begin{table}[t]
		\caption{Tracking performance comparison.}
		\centering
		\begin{tabular}{lcc}
			\hline
			Method & Position RMSE [m] & Orientation RMSE [rad] \\
			\hline
			PD + ZOH      & 0.003786 & 0.012861 \\
			PD + FOH      & 0.005727 & 0.017894 \\
			PD + LPF      & \textbf{0.003660} & \textbf{0.012496} \\
			Passive + ZOH & 0.008076 & 0.034089 \\
			Passive + FOH & 0.009899 & 0.037536 \\
			Passive + LPF & \textbf{0.007946} & \textbf{0.033748} \\
			\hline
		\end{tabular}
		\label{tab:tracking_rmse}
	\end{table}
	
	\subsection{Control Effort Analysis}
	
\begin{figure}
		\centering
		\includegraphics[width=3.45in]{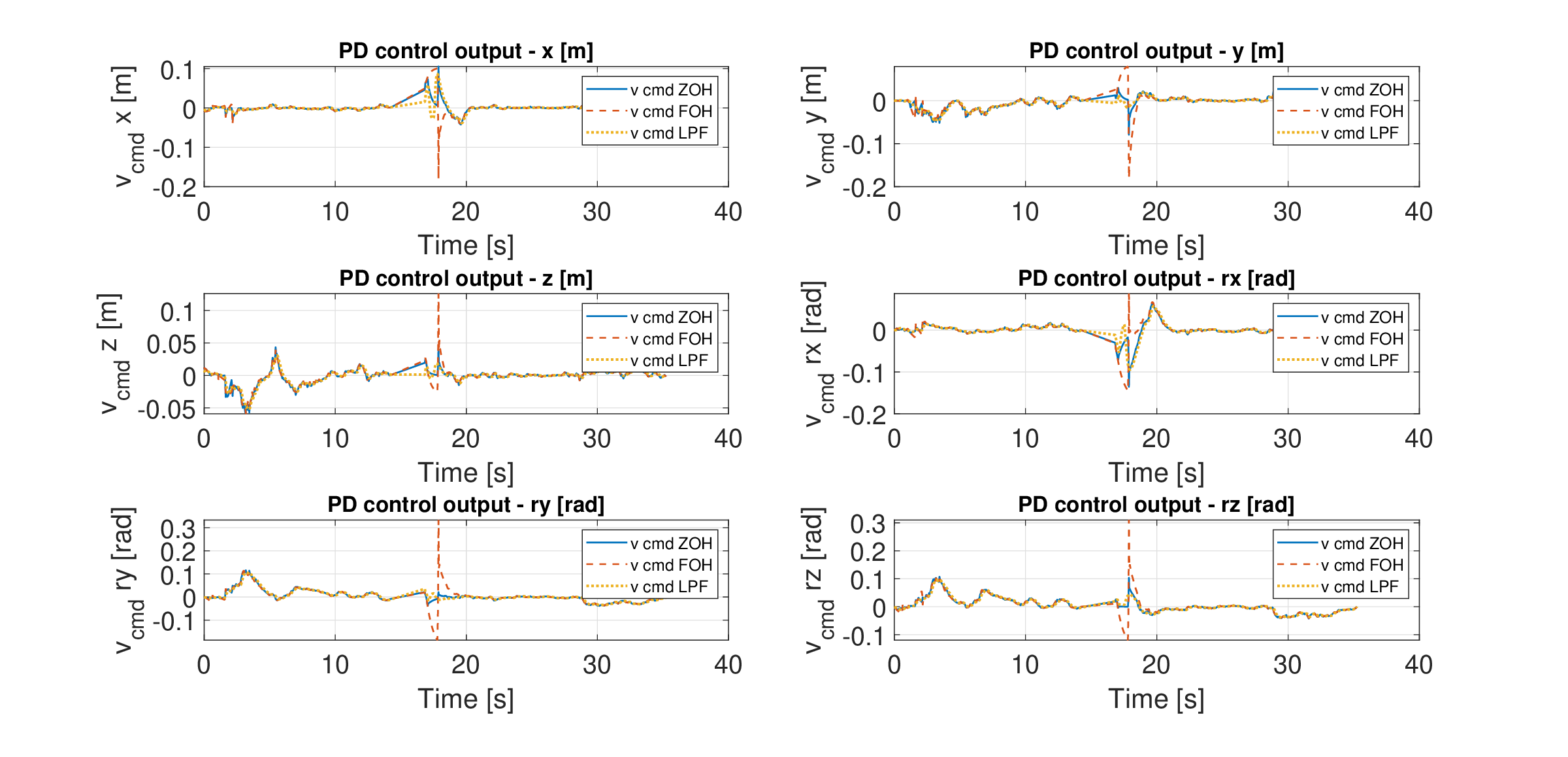}
		\caption{Control outputs generated by the PD controller using the ZOH, FOH, and LPF reconstruction strategies.}
		\label{fig:pd_control_effort}
	\end{figure}

\begin{figure}
		\centering
		\includegraphics[width=3.45in]{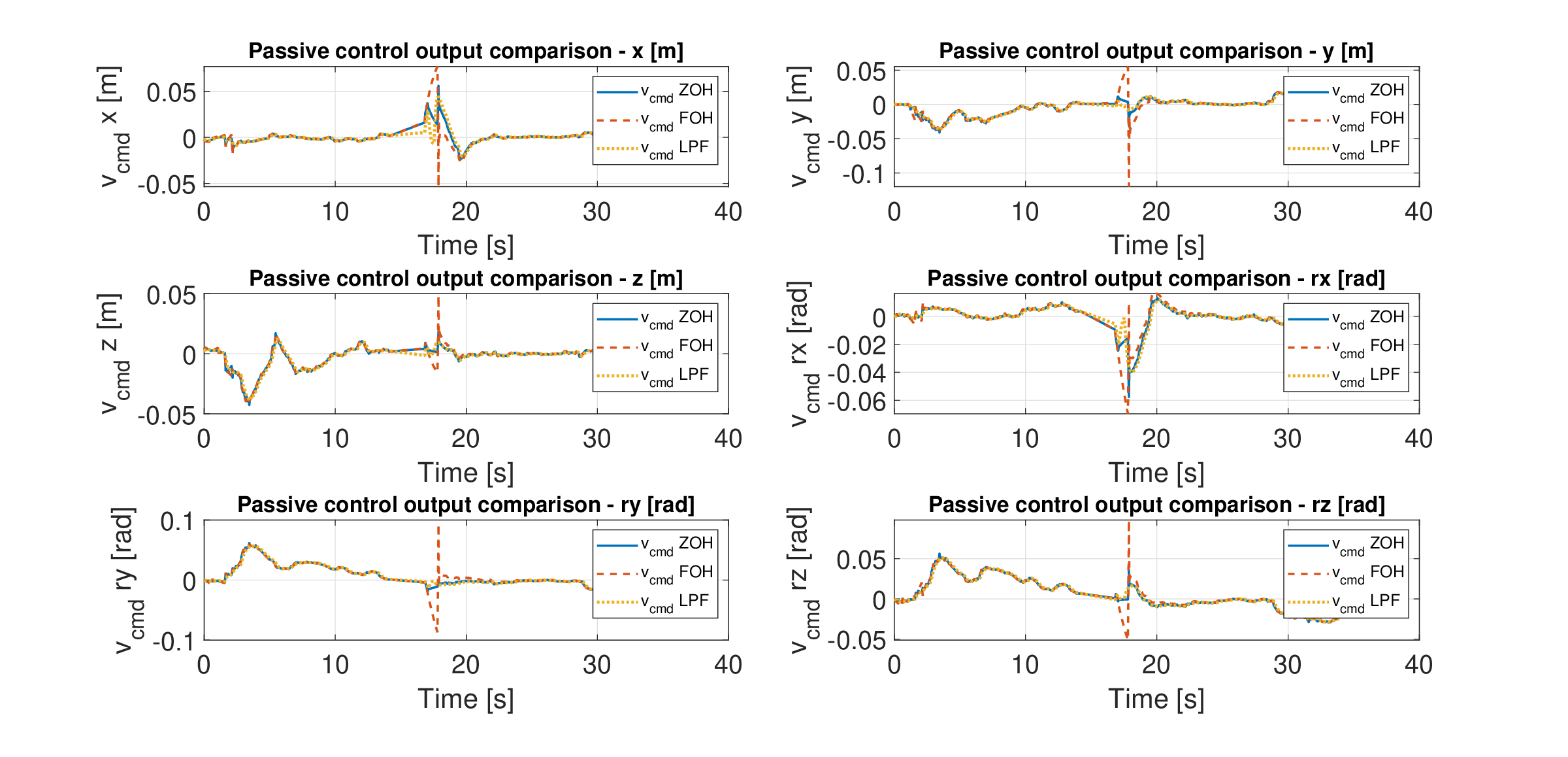}
		\caption{Control outputs generated by the passive synchronization controller using the ZOH, FOH, and LPF reconstruction strategies.}
		\label{fig:Passive_vcmd_comparison}
	\end{figure}

	Besides tracking accuracy, the control effort generated by each controller was analyzed. The commanded translational and rotational velocities produced by the PD and passive synchronization controllers are shown in Figs.~\ref{fig:pd_control_effort} and \ref{fig:Passive_vcmd_comparison}.
	
	For both controllers, the control commands remain smooth during nominal communication conditions. The largest control transients occur around $t\approx17$~s, corresponding to the communication disturbance identified during the network characterization stage, where irregular packet arrivals temporarily affect the reconstructed reference.
	
	Among the three reconstruction strategies, FOH produces the largest control peaks for both translational and rotational velocities. Although FOH guarantees signal continuity through linear interpolation, the reconstructed reference exhibits pronounced transient variations during the communication disturbance, which require larger corrective actions from the controller.
	
	The ZOH reconstruction generates moderate control peaks associated with its piecewise-constant reference updates. While these discontinuities introduce abrupt command changes, the resulting control effort remains lower than that produced by FOH.
	
	LPF consistently provides the smoothest control commands. By attenuating the high-frequency components introduced by communication jitter, the low-pass filter significantly reduces the amplitude of the control peaks while preserving the overall trajectory evolution. This behavior is particularly noticeable in the translational $x$ component and in the rotational variables $rx$ and $rz$, where LPF effectively limits the transient control actions generated during the communication disturbance.
	
	A comparison between Figs.~\ref{fig:pd_control_effort} and \ref{fig:Passive_vcmd_comparison} further shows that the passive synchronization controller generally produces smaller control commands than the classical PD controller under identical communication conditions. The passive formulation naturally introduces a more dissipative closed-loop behavior, reducing aggressive corrective actions while maintaining stable synchronization.
	
	\begin{table}[t]
		\caption{Control-effort comparison.}
		\centering
		\begin{tabular}{lcccc}
			\hline
			Method &
			Pos. RMS &
			Pos. Peak &
			Ori. RMS &
			Ori. Peak \\
			\hline
			PD + ZOH      & 0.023083 & 0.138008 & 0.041530 & 0.171608 \\
			PD + FOH      & 0.034254 & 0.283982 & 0.058798 & 0.463577 \\
			PD + LPF      & \textbf{0.022286} & \textbf{0.089770} & \textbf{0.040275} & \textbf{0.144245} \\
			Passive + ZOH & 0.017629 & 0.066756 & 0.029128 & 0.083710 \\
			Passive + FOH & 0.022104 & 0.139905 & 0.032611 & 0.135733 \\
			Passive + LPF & \textbf{0.017329} & \textbf{0.053941} & \textbf{0.028801} & \textbf{0.076730} \\
			\hline
		\end{tabular}
		\label{tab:control_effort}
	\end{table}
	
	Tables~\ref{tab:tracking_rmse} and \ref{tab:control_effort} summarize the quantitative performance obtained under the different reconstruction strategies.
	
	The tracking results show that the classical PD controller achieves lower RMSE values than the passive synchronization controller for all reconstruction methods. The lowest tracking error is obtained with the LPF reconstruction, reaching a position RMSE of 0.003660~m and an orientation RMSE of 0.012496~rad.
	
	In contrast, the passive synchronization controller consistently requires less control effort than the classical PD controller. For every reconstruction strategy, both the RMS and peak values of the commanded translational and rotational velocities are significantly reduced. This reduction is particularly evident for the LPF reconstruction, where the position RMS decreases from 0.022286 to 0.017329 and the orientation peak decreases from 0.144245 to 0.076730.
	
	Regarding the reconstruction methods, FOH produces the largest control peaks despite providing continuous reference trajectories. The linear interpolation amplifies the transient response during the communication disturbance, resulting in considerably higher control actions than those obtained with ZOH or LPF. Conversely, LPF consistently achieves the smallest control effort while preserving tracking accuracy, making it the most suitable reconstruction strategy among the evaluated methods.
	
	Overall, the experimental results reveal a clear trade-off between tracking accuracy and control effort. The classical PD controller provides the highest tracking accuracy, whereas the passive synchronization controller generates smoother control actions with significantly lower actuation demands. These results demonstrate that combining passivity-based synchronization with LPF reconstruction provides an attractive solution for cloud-based unilateral teleoperation operating under realistic Internet communication uncertainties.
	
	
	\section{Conclusion}
	The experimental results demonstrated that the reference reconstruction strategy influences the teleoperation performance. LPF achieved the best overall performance, with the lowest tracking errors and the smallest control effort. FOH generated the largest transient control peaks during communication disturbances, indicating that linear interpolation may amplify abrupt variations under high-frequency packet arrivals. ZOH remained the simplest and most robust reconstruction strategy, but its piecewise-constant behavior introduced discontinuities that increased control activity.
	
	The comparison between the two controllers showed that the classical PD controller achieved slightly lower tracking errors, whereas the passive synchronization controller consistently reduced the commanded control effort while preserving stable operation under the same communication conditions. These results indicate that the passive synchronization framework can improve motion smoothness without compromising closed-loop stability, which makes it particularly attractive for teleoperated robotic applications such as robotic haircutting.
	
	The proposed framework provides an experimentally validated benchmark for evaluating delay-compensation and passivity-based control strategies under realistic cloud communication conditions. The obtained results highlight the importance of combining appropriate reference reconstruction with robust control algorithms for Internet-based teleoperation systems.
	
	Future work will investigate predictive delay-compensation techniques, adaptive passivity-based synchronization methods, and experimental validation with packet losses, larger communication uncertainties, and bilateral teleoperation architectures.
	
	\ifCLASSOPTIONcaptionsoff
	\newpage
	\fi

	\bibliographystyle{IEEEtran}
	\bibliography{IEEEabrv,Bibliography}

	\vfill
	
\end{document}